\documentclass[conference]{IEEEtran}

\usepackage{cite}
\usepackage{amsmath,amssymb,amsfonts}
\usepackage{algorithmic}
\usepackage{graphicx}
\usepackage{textcomp}
\usepackage{color}
\usepackage{array}
\usepackage{cite}

\usepackage{xcolor}
\usepackage{array}
\newcolumntype{P}[1]{>{\centering\arraybackslash}m{#1}}
\usepackage{url} 

\def\BibTeX{{\rm B\kern-.05em{\sc i\kern-.025em b}\kern-.08em
    T\kern-.1667em\lower.7ex\hbox{E}\kern-.125emX}}

\begin{document}
\title{Automated Segmentation and Recurrence Risk Prediction of Surgically Resected Lung Tumors with Adaptive Convolutional Neural Networks}
\author{Marguerite B. Basta$^{1}$, Sarfaraz Hussein$^{2}$, Hsiang Hsu$^{1}$, and Flavio P. Calmon$^{1}$ \bigskip\\

\textit{$^1$John A. Paulson School of Engineering and Applied Sciences, Harvard University}\\
\textit{$^2$Center for Research in Computer Vision (CRCV), University of Central Florida}
}

\maketitle

\begin{abstract}
Lung cancer is the leading cause of cancer related
mortality by a significant margin. While new technologies, such as
image segmentation, have been paramount to improved detection
and earlier diagnoses, there are still significant challenges in
treating the disease. In particular, despite an increased number
of curative resections, many postoperative patients still develop
recurrent lesions. Consequently, there is a significant need for
prognostic tools that can more accurately predict a patient’s risk
for recurrence.

In this paper, we explore the use of convolutional neural
networks (CNNs) for the segmentation and recurrence risk
prediction of lung tumors that are present in preoperative
computed tomography (CT) images. First, expanding upon recent
progress in medical image segmentation, a residual U-Net is used
to localize and characterize each nodule. Then, the
identified tumors are passed to a second CNN for recurrence risk
prediction. The system’s final results are produced with a random
forest classifier that synthesizes the predictions of the second
network with clinical attributes. The segmentation stage uses the LIDC-IDRI dataset and achieves a dice score of 70.3\%. The recurrence risk stage uses the NLST dataset from the National Cancer institute and achieves an AUC of 73.0\%. Our proposed
framework demonstrates that first, automated nodule segmentation
methods can generalize to enable pipelines for a wide range
of multitask systems and second, that deep learning and image
processing have the potential to improve current prognostic tools.
To the best of our knowledge, it is the
first fully automated segmentation and recurrence risk prediction
system.

\end{abstract}


\label{sec:introduction}
\section{Introduction}
\begin{figure}[!tb]
    \includegraphics[width=.5\textwidth]{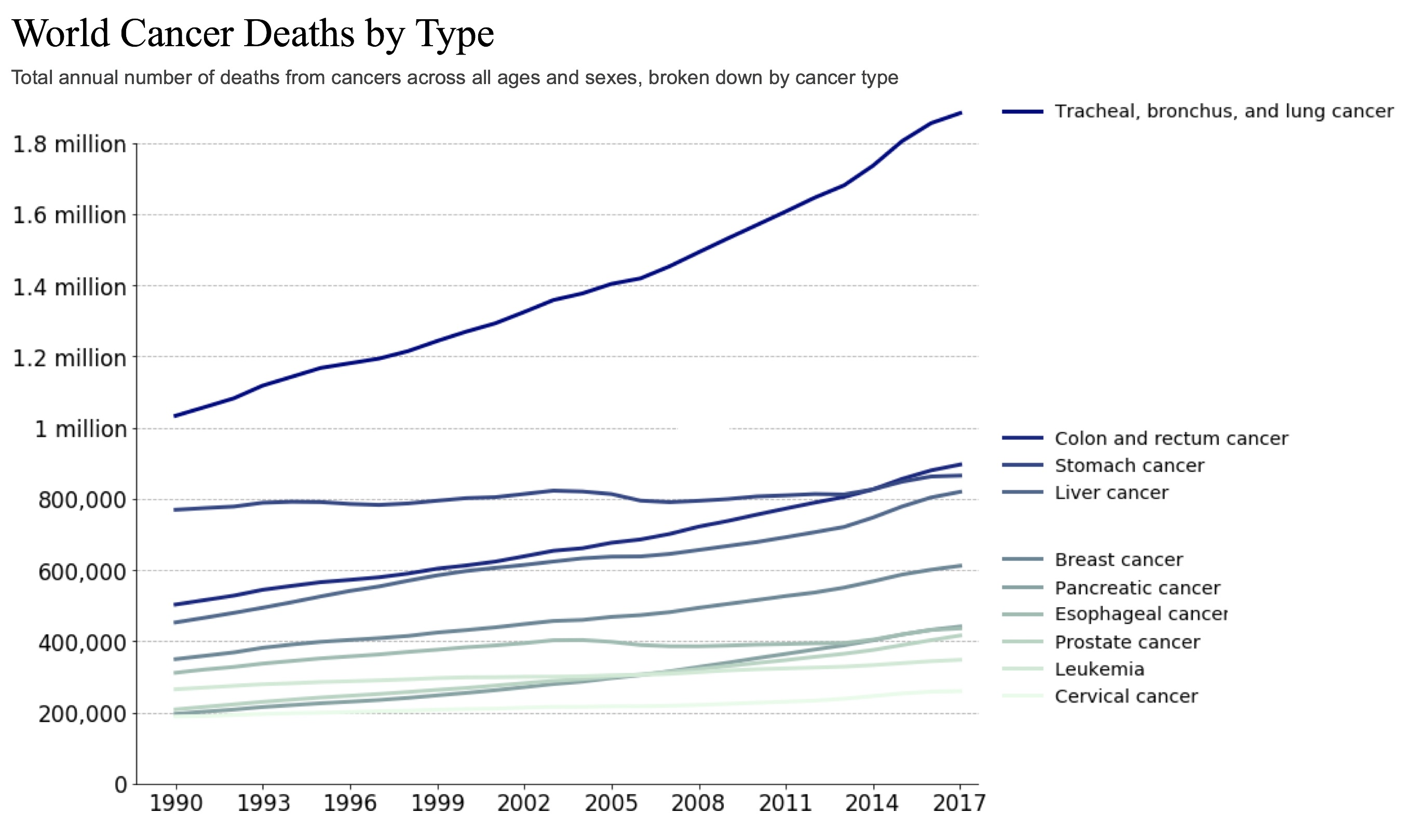}%
    \caption[World lung cancer deaths.]{Total number of world deaths from 1990-2017 caused by a range of different cancer types. Tracheal, bronchus, and lung cancers consistently remain the most deadly causing 1.9 million deaths in 2017.  The data for the plot was collected by the Institute of Health Metrics and Evaluation (IHME) as part of the Global Burden of Disease (GBD) project and was organized by \cite{owidcancer}.}
    \label{fig:CancerDeaths}
\end{figure}

Lung cancer causes significantly more deaths each year than any other form of cancer. In fact, even though there has been significant progress in the diagnosis and treatment of the disease, it is still projected to remain the leading cause of cancer related mortality through 2030  \cite{Rahib2014} (see Figure \ref{fig:CancerDeaths}). While improved medical technology can enable earlier detection and therefore increase the number of curative tumor resections, over 30\% of postoperative patients will still develop recurrent lesions \cite{Consonni2015}. On average, these patients do not survive more than a year \cite{Sugimura2007}. Consequently, disease recurrence contributes significantly to lung cancer's startlingly low 5 year survival rate of 4-17\% depending on stage and geographic location \cite{Hirsch2017}. 

With the rise of precision medicine and continued development of new treatment options however, there has been an increased push for better and more individualized recurrence prognostication tools \cite{hayes2014personalized}. The failure to accurately evaluate risk for recurrence in the past is largely due to the inadequacy of current methods, particularly the Tumor, Node, Metastasis (TNM) staging system.\footnote{T (tumor) refers to the extent of the primary tumor. N (node) refers to the extent of cancer that has spread to nearby lymph nodes. M (metastasis) refers to whether the cancer has spread to distant parts of the body.} The system is the medical standard for defining the extent and spread of cancer and is used to break patients up into 5 numerical stages (stage 0 to stage IV) \cite{CancerStagingNCI}. Despite the fact that it lacks the granularity required to accurately predict risk of recurrence on an individual basis, in clinical settings, it is still considered the best tool for the task \cite{Osarogiagbon2012}. Specifically, post-surgical recurrence rates for stage I patients have been reported to range from 27\% to 38\% \cite{Hung2009}. However, with the TNM system, stratifying a patient as high or low risk for recurrence and the subsequent treatment decisions (like whether or not to administer adjuvent therapy), is usually determined by whether or not their cancer has progressed to stage II \cite{Subramanian2010}.\footnote{Adjuvent therapy is treatment, usually chemotherapy, that is administered in addition to surgery to reduce the risk of recurrent disease \cite{Tsuboi2007}.} Thus, there is a persistent need for a more granular prognostic tool, and in particular, for one that can more accurately identify high risk stage I patients or low risk stage II patients \cite{Subramanian2010}. Fortunately, new medical technologies are continuing to enable the development of potentially transformative methods \cite{JungKoo2017, Hosny2018}. 

Recently, the search for a better recurrence prognostication tool has found its way to \emph{radiomics} -- an emerging field of research that aims to extract high-dimensional data from radiographic medical images \cite{Rizzo2018, Lambin2012}. While the use of the field for recurrence prediction is novel, the few studies that have utilized these methods have had promising results \cite{JungKoo2017, Hosny2018}. Nonetheless, both \cite{JungKoo2017, Hosny2018} rely on radiologists to localize the tumors by hand and do not have a fully automated pipelines. As a result, extending the methodologies in \cite{JungKoo2017, Hosny2018} to unannotated datasets is intractable if a radiologist cannot be accessed. Unfortunately, to the best of our knowledge, there are no published lung imaging datasets that contain both the information to determine recurrence status (as opposed to just survival time) and prerecorded annotations from radiologists. Nonetheless, although relatively untested prior to this study, a potential alternative is to localize and geometrically characterize tumors from datasets without annotations using automated image segmentation. 

In recent years, significant research effort has gone into developing automated methods for segmenting lung tumors in order to aide the radiology process. While certain semi-automated methods have enabled software that facilitates radiologist led segmentation, there is still a need for full automation. This has become especially true due to the current state of lung cancer screening. In the past decade, significant evidence has shown that using low-dose CT scanning as opposed to chest radiography (i.e. X-ray) for screening reduces lung cancer mortality rates \cite{Aberle2011}. These findings have led to recommendations from the American College of Radiography \cite{Kazerooni2015}, the US Preventive Services Task Force \cite{Moyer2014}, and the National Comprehensive Cancer Network \cite{Wood2018} reemphasizing that CT screens should be the first step in managing lung abnormalities. As a result, CT screening has become the default modality in lung cancer imaging. However, the manual analysis of a CT scan, which will usually contain hundreds of slices, is an extremely laborious process. Not only does this lead to less efficient clinical protocols, but it also makes building large scale, manually segmented datasets extremely difficult. However, as we mentioned previously, the issues with manually segmenting datasets can potentially be resolved by the development and integration of automated nodule segmentation. Fortunately, with the 2011 release of the LIDC-IDRI dataset \cite{Armato2011}, a large-scale lung cancer imaging collection that is both publically available and analyzed by multiple radiologists, research on automated nodule segmentation has accelerated rapidly \cite{Cavalcanti2016, Wang2017, Wang2019, Nithila2016}. Still, few of these algorithms have demonstrated the capability to extend past LIDC-IDRI or been used in automated, multitask pipelines. In this study, we explore this relatively untested area with recurrence risk prediction. 

\begin{figure}[!tb]
    \includegraphics[width=.5\textwidth]{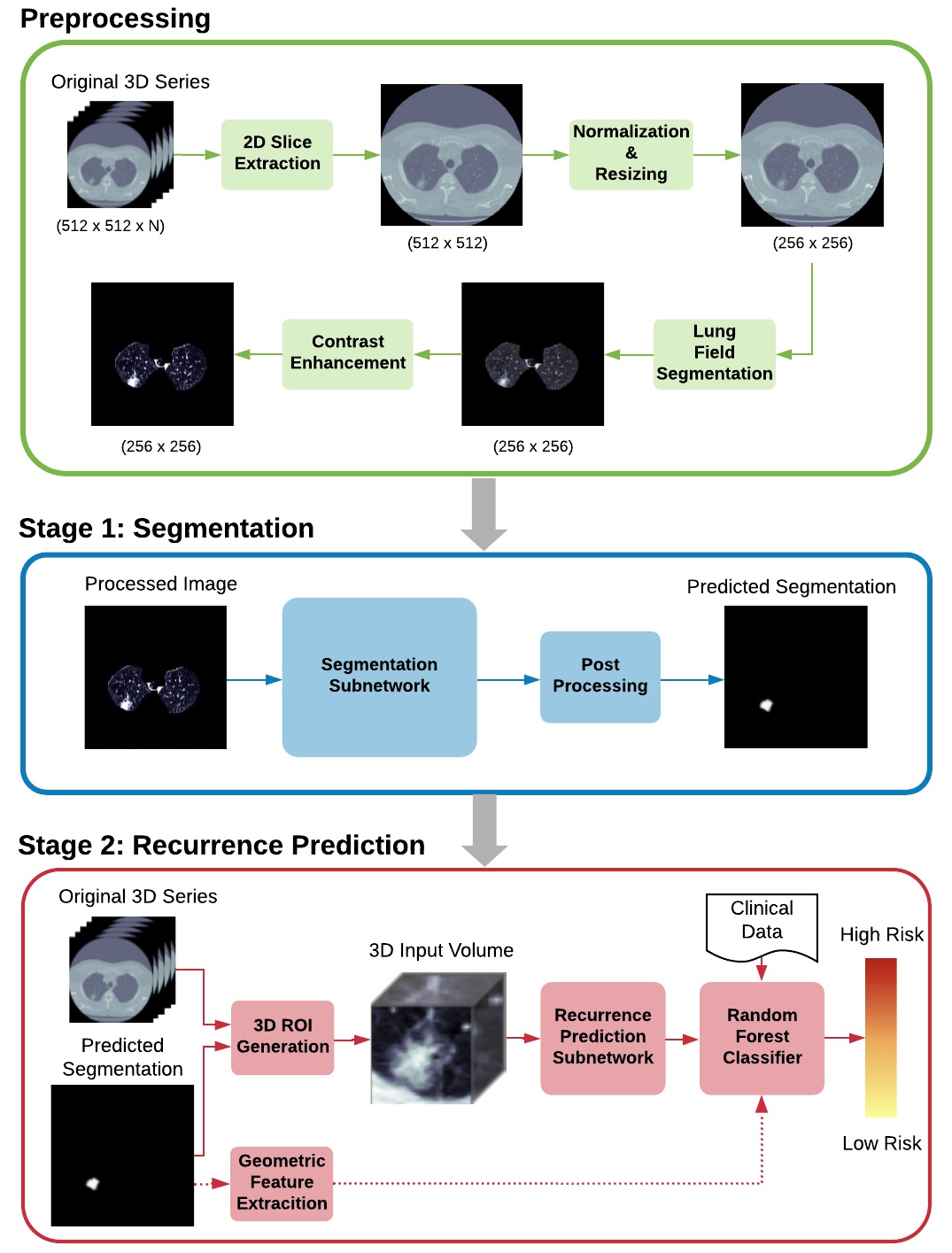}%
    \caption[Proposed Nodule Segmentation and Recurrence Prediction Framework]{Proposed nodule segmentation and recurrence prediction framework. First, the primary 2D slice (slice containing the largest tumor diameter) is preprocessed to resize and normalize the image, isolate lung fields, and increase contrast. It is then fed into a segmentation network where a tumor mask is predicted and run through post processing. Next, the predicted mask is used to retrieve a 3D ROI from the original CT series, which is fed into a 3D CNN for recurrence prediction. Finally, the CNN's output, clinical data, and geometric features of the tumor, which are extracted from the predicted segmentation, are used as features for a random forest classifier.}
    \label{fig:pipeline_overview}
\end{figure}

The system we propose consists of preprocessing and two main prediction stages. In the first stage, a 2D axial slice containing nodule tissue is taken from a 3D CT series, and sent as input to a segmentation subnetwork. The predicted segmentation is then post-processed and passed to the second stage to generate a 3D region of interest (ROI) around the nodule. The ROI is sent as input into the second subnetwork for recurrence risk prediction and assigned a risk score between 0 and 1. A cutoff of 0.5 is also applied to evaluate the system's performance as a binary classifier. In the last part of our study, we build a model that synthesizes the predictions of our network with clinical data and geometric tumor features that are extracted from our predicted segmentations. We implement this final model as a random forest classifier \cite{liaw2002classification}. An illustration of the entire system's framework is provided in Figure \ref{fig:pipeline_overview}.
\footnote{All code used to implement the system and run experiments is available in this project's online repository: \url{https://github.com/maggiebasta/lung-cancer-thesis}.}

Ultimately, the successful implementation of nodule segmentation algorithm within a multitask, multi-dataset pipeline exhibits the power of current medical image segmentation. To the best of our knowledge, our proposed framework is the first fully automated segmentation and classification system for recurrence risk prediction. Our main contributions are summarized below: 
\begin{enumerate}
    \item We demonstrate that nodule segmentation algorithms can generalize to other independent datasets.
    \item We show that these algorithms can be integrated into classification pipelines that are trained using these different datasets.
    \item We make the NLST dataset viable for recurrence prediction via automated segmentation. This enables algorithms to use the extensive of features included in NLST even when a radiologist is not accessible. 
\end{enumerate}

\section{Materials and Preprossessing}
In the following section, we discuss the datasets used and the initial pipelinefor processing their images.  To start, we provide an overview of each of the datasets,their collection methods, and their respective roles within our framework

\subsection{Datasets}

We utilize two independent datasets for our study: 
\begin{itemize}
    \item \textbf{LIDC-IDRI} \cite{Armato2011}: This dataset is used to train the segmentation stage of our system. The original dataset before cleaning consists of 1,018 low-dose CT image scans from 1,010 patients in DICOM format, each with a varying number of cross-sectional 2D slices. Additionally, each of the series has an associated XML file containing the annotations of 4 separate radiologists. 
    \item \textbf{NLST} \cite{nlst}: We use a subset of the NLST CT image collection to train and test our recurrence prediction stage. The subset provided contains only patients with confirmed lung cancer. Before cleaning, it consists of 1,165 sets of low-dose CT scans from 622 patients in DICOM format, each with a varying number of cross-sectional 2D slices. In addition to the images, there is a significant amount of clinical, treatment, and outcome data included.
\end{itemize}

\noindent A summary of the important attributes of each dataset (after cleaning and feature engineering) is provided in Table \ref{table:dataset_summary}.

\subsection{Data Cleaning and Feature Engineering}

In order to remove outliers and inconsistencies before training, both the LIDC-IDRI and NSLT datasets go through a data cleaning process. For LIDC-IDRI, any scans or nodules that match one or more of the following criteria is excluded from our study: scans with large slice thickness ($>$ 2.5mm), scans with missing slices, or inconsistent spacing, nodules with diameters $<$ 3mm (marked in the dataset by a single point rather than an entire contour, nodules not marked by the majority of radiologists (at least 3). For NLST, cleaning is done on a per patient and per nodule basis. Patients and nodules excluded are: patients who had non-surgical primary treatments, like chemotherapy or radiation therapy, patients who had residual disease left after their surgery, patients who did not remain in contact until the conclusion of the study, patients who have multiple, but not identical, scans included for the same year. (The dataset does not indicate which of these scans corresponds to the reported tumors), nodules with diameter $<$ 4mm. (Nodule diameter is not recorded in NLST if it is less than 4mm). \\

Following data cleaning, the substantial amount of clinical data present in NLST allows us to infer additional features. Most prominently, cancer recurrence (our target outcome variable) is inferred from the progression data and treatment data. If the patient had a cancer progression, the time of the progression is compared to the time of surgery. Additionally, we infer an adjuvant therapy feature. If the patient has any documented chemotherapy treatments in addition to surgery, we compare the time the treatment is administered to the time of surgery. If it is determined that chemotherapy was administered post-surgically, the patient is marked as having received adjuvant therapy.\\

\begin{table}[tb!]
\begin{center}
\caption[Dataset Summary]{\textbf{Dataset summary}\\}
\label{table:dataset_summary}
\begin{tabular}{||P {1cm}|P {1.75cm}| P {3.75cm}||} 
 \hline
 Dataset & Purpose & Important Attributes\\ [0.5ex] 
 \hline\hline
 LIDC-IDRI & segmentation training and testing & radiologists' annotations, primary axial slices\\ 
 \hline
 NLST & recurrence risk prediction training and testing & disease progression, residual disease after surgery, treatments administered, cause of death, cancer stage, demographic information, adjuvant therapy, primary axial slices\\
 \hline
\end{tabular}
\end{center}
\end{table}
%

\section{Lung Nodule Segmentation}

\subsection{Segmentation Architecture}
\begin{figure*}[!tb]
  \centering\includegraphics[width=.75\textwidth]{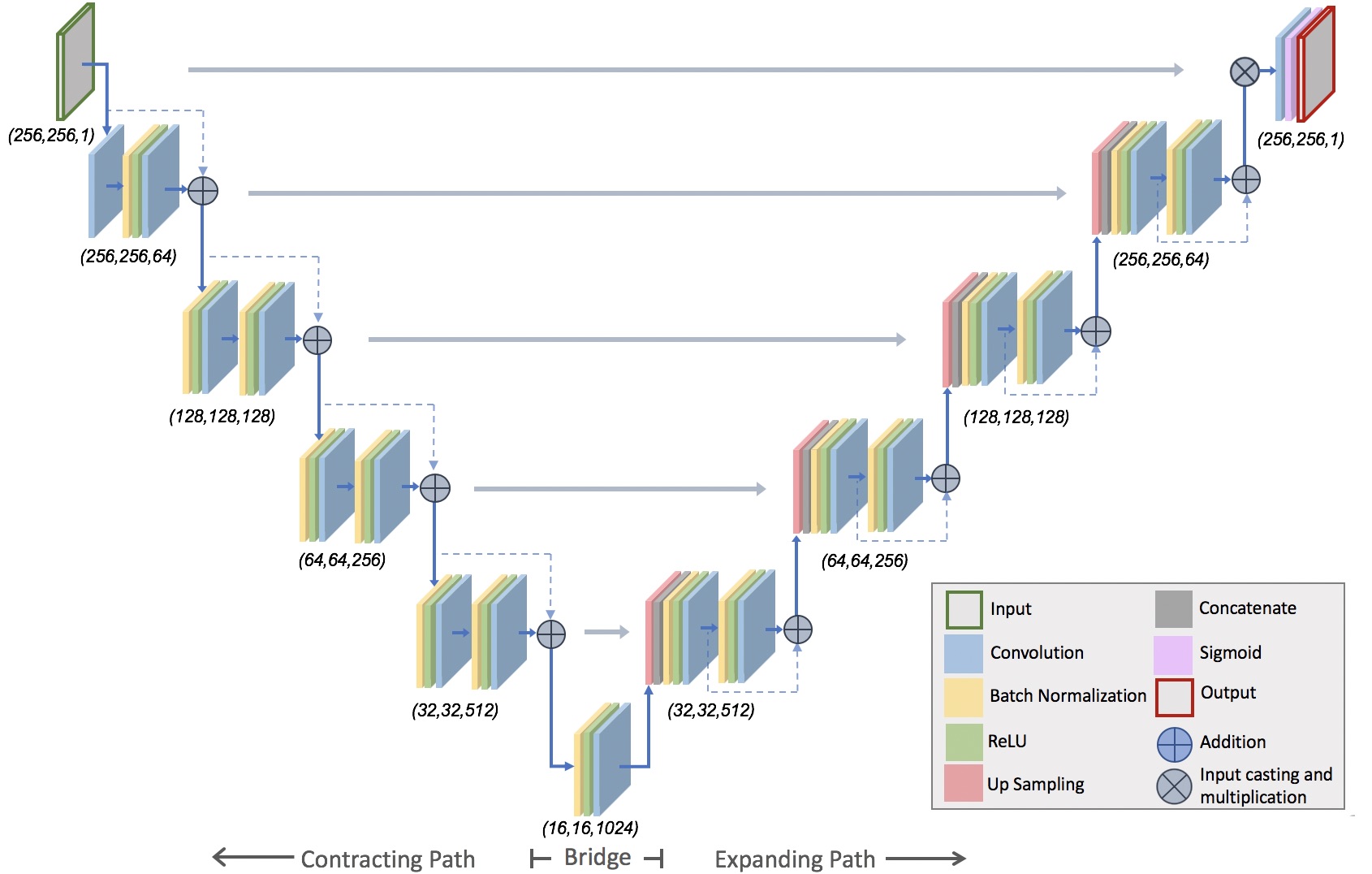}%
    \caption[Architecture of the Proposed Residual U-Net for Segmentation]{Architecture of the proposed residual U-Net for segmentation. The output shape, (image $x$ size, image $y$ size, and number of channels), is denoted below each unit. Blue arrows represent the main path of the network, dotted arrows represent residual connections, and grey arrows represent the connection between the mirrored units of the contracting and expanding paths. The colors of the layers represent different operations.}
    \label{fig:UnetArchitecture_2}
\end{figure*}

For our lung nodule segmentation model, we choose to use an adapted U-Net \cite{Ronneberger}. The U-Net, a CNN variant, has recently become the state of the art for many segmentation tasks. The specific architecture we use, which is strongly inspired by \cite{Zhang}, adjusts the original architecture to contain residual connections.\footnote{Residual connections (first proposed by \cite{He}) are constructed by adding the unit's original input to its output.}  An illustration is provided in \ref{fig:UnetArchitecture_2}. It can be best understood by inspecting the individual units within the three main parts -- the contracting path, the bridge, and expanding path.

\begin{enumerate}
    \item  \textbf{Contracting Units}:
    Similar to \cite{Zhang}, each residual unit within the contracting path applies a batch normalization and ReLU activation to the previous unit's output. This is followed by a $(3\times3)$ convolution with a stride length of 2. The process is then repeated a second time, but with a convolution stride length of 1. Finally, the original input is added to the output of the second convolutional layer and passed on to the next unit. Unlike the original U-Net, the stride length of 2 in the first convolution serves as the downsampling operation, so no max pooling operation is applied. 
    \item \textbf{Bridge Unit}:
    Next, the bridge unit consists of this same batch normalization, activation, convolution sequence. However, the sequence is applied only once before the output is passed to the expanding path. Also, there is no residual connection across the unit.
    \item \textbf{Expanding Units}:
    The units in the expanding path have a similar structure to the contracting units. However, first, a $(2\times 2)$ up-sampling operation and concatenation with the output of the mirrored contracting unit is applied. Also, both convolutional layers in the expanding path have stride lengths of 1. 
\end{enumerate}
Following the expanding path, the image is passed through a $(1\times1)$ convolution and sigmoid activation. This activation is then followed by the application of a binary mask to eliminate any false positives detected outside of the lung region. The mask is generated from the network's original input. Specifically, since our preprocessing routine yields a normalized image with the background removed, applying a ceiling operation to this image creates a binary mask of the lung field. Then, applying the mask to the output of the sigmoid removes any falsely activated pixels from outside of the lung region.

\subsection{Segmentation Training}

Following the preprocessing routine discussed in the previous section, the images are partitioned randomly into training, testing, and validation sets based off patient ids. In order to maximize the number of images for training, all slices containing nodule tissue are included in the training and validation sets. For the test set however, only the image with the largest cross-sectional tumor diameter is considered. (During recurrence prediction in the next stage of the system, only slices with the largest cross-sectional tumor diameter are used to segment each nodule).

The model is implemented using the Keras 2.2.4 using tensorflow 1.14.0 as the backend. Due to the significant imbalance of non-nodule to nodule pixels, weighted binary cross-entropy is used as the loss function: 
\begin{equation}
  \label{weighted_bce}
  L_{wbce} = -\frac{1}{N}\sum_{i=1}^{N} w\cdot y_i \cdot \log(h_{\theta}(x_i)) + (1-y_i) \cdot \log(1 - h_{\theta}(x_i))
\end{equation}
where $N$ is the number of samples, $w$ is the weight given to the positive class, $y_i$ is the true label for the $i$'th sample, $h_\theta$ is the network with weights $\theta$, and $x_i$ is the input for the $i$'th sample. We use a weight of 12.0 for the positive class. For our optimizer, we use the Adam optimizer with a global learning rate of 0.0001.

Another critical element of our training process is data augmentation.
Due to the limited number of samples and the similarity between images from the same
patient, augmentation is crucial to avoid overfitting. All augmentation is done in real time during training instead of as part of preprocessing. The operations include random brightness shifts, rotations, shears, zooms, and horizontal flips. The magnitude of each of the operations is chosen randomly during training, and all images are also normalized to the 0-1 range at the end of the augmentation process.


\subsection{Segmentation Post-Processing}
After the final output from the network, a small amount of post-processing is done. For certain predicted segmentations, there are multiple activated regions. This can be due to the presence of more than one nodule in the same image, or false positives. In order to both simplify the transition to the next stage of the system (recurrence classification on a single nodule) and eliminate potential false positives, we select only one of these regions to remain activated in the final segmentation. The region selected it the most \textit{densely} activated region.

\subsection{Segmentation Results and Experiments}
\label{eval_metrics}
We evaluate our model using several metrics that compare the predicted segmentations with the manual segmentations made by the radiologists:\footnote{ We choose not to measure accuracy as it becomes extremely inflated as a result of the class imbalance.}
\begin{enumerate}
    \item\textit{Dice coefficient}: a measurement of the relative overlap between two images.
    \begin{equation}
      \label{dice}
      Dice(X, Y) = \frac{2(|X|\cap|Y|)}{|X|+|Y|)}
    \end{equation}

    where $\cap$ is the intersection operator and $X$ and $Y$ are the two images in question. A dice coefficient of 1 occurs when there is a perfect overlap. A dice coefficient of 0 occurs when there is no overlap. 
    \item \textit{Precision}
    \item \textit{Recall}
    \noindent
\end{enumerate}

To test the performance of our model, we compare it with two other architectures trained within the same pipeline. The architectures chosen are the original U-Net \cite{Ronneberger} and the original ResUnet \cite{Zhang}. Table \ref{table:segmentation_results_pre} shows a comparison of the results on the test set before post-processing, and Table \ref{table:segmentation_results_post} shows a comparison of the results on the test set after post-processing. \\

\begin{table}[tb!]
\caption[Segmentation Results Before Post-Processing]{Segmentation results before post-processing. \\}
\label{table:segmentation_results_pre}
\centering
\begin{tabular}{||P {3cm}|P {1.5cm}| P {1cm}| P {1cm}||} 
 \hline
 Architecture & Dice Coefficient & Recall & Precision \\ [0.5ex] 
 \hline\hline
 Proposed Model & 67.7\%   & 81\% & 64\%\\ 
 \hline
 Original ResUnet & 64.2\% & 76\% & 66\%
 \\
 \hline
 Original U-Net & 50.9\% & 92\% & 38\%\\
 \hline
\end{tabular}
\end{table}

\begin{table}[tb!]
\caption[Segmentation Results After Post-Processing]{Segmentation results after post-processing.\\}
\label{table:segmentation_results_post}
\centering
\begin{tabular}{||P {3cm}|P {1.5cm}| P {1cm}| P {1cm}||} 
 \hline
Architecture & Dice Coefficient & Recall & Precision\\ [0.5ex] 
 \hline\hline
 Proposed Model & 70.3\%   & 78\% & 69\%\\ 
 \hline
 Original ResUnet & 64.4\% & 73\% & 68\% \\
 \hline
 Original U-Net & 54.3\% & 76\% & 38\%\\
 \hline
\end{tabular}
\end{table}

We also benchmark our approach against the recent lung nodule segmentation literature.  A wide variety of models are used for this comparison. Specifically, approaches include a combined active contour and Fuzzy C-means model \cite{Nithila2016}, a region based CNN \cite{Wang2019}, and a recurrent 3D U-Net \cite{Kamal}. The results are shown in table \ref{table:other_segmentation_comparison}.\footnote{Other studies have reported dice scores as high as 82\% but reduce the problem space to a cropped region around the nodule \cite{Wang2017}. Therefore, we do not consider them as a benchmark.} With the exception of the Recurrent 3D U-Net proposed by Kamal et al. \cite{Kamal}, our model outperforms the other methods. It is unclear if the use of different datasets has a significant effect on these results (\cite{Kamal} is the only study not using LIDC-IDRI), but our approach still proves to be competitive.

\begin{table}[tb!]
\caption[Comparison With Other Segmentation Models]{Comparison with other segmentation models.}
\label{table:other_segmentation_comparison}
\centering
\begin{tabular}{||P {1cm}|P {2.5cm}|P {1.7cm}| P {1.25cm}||} 
 \hline
Study & Approach & Dataset & Dice Coefficient\\ [0.5ex] 
 \hline\hline
Proposed Model & Adapted Residual U-Net & LIDC-IDRI & 70.3\%\\ 
 \hline
Kamal et al. \cite{Kamal} & Recurrent 3D U-Net & IEEE 2018 VIP Cup & 74.0\%\\
 \hline
Nithila et al. \cite{Nithila2016} & Region Based CNN & LIDC-IDRI & 60.0\%\\
 \hline
Wang et al. \cite{Wang2019} & Active contour and Fuzzy C-mean & LIDC-IDRI & 60.0\%\\
 \hline
\end{tabular}
\end{table}

\subsection{Segmentation Discussion}
By taking advantage of methods from both inside \cite{Ronneberger} and outside \cite{Zhang} the medical community, we are able to implement a model for lung nodule segmentation that is competitive with the current state of the art. Ultimately, however, for our purposes, the most significant way to validate our model is to demonstrate its efficacy in our full pipeline. This is discussed in the following section, where we present a recurrence risk prediction system that generates its input from our model's predicted segmentations. 

The recurrence prediction framework we present consists of several intermediate steps. First, the predicted segmentation from the previous stage is used to generate a 3D ROI around the identified tumor. Next, this ROI is passed as input to a recurrence risk prediction CNN. Finally, the CNN's output, clinical data, and geometric features of the tumor, which are extracted from the previous stage's predicted segmentation, are used as features in a random forest classifier. Each of these steps is discussed in further detail below.

\section{Recurrence Prediction}
\subsection{Recurrence Prediction ROI Generation}

The first step in ROI generation is converting from voxels to world coordinates (mm). That is, the raw 3D volume is resampled so that each data point represents a $1\times 1\times 1$ mm cube of real world data. 
After resampling, a ROI can be extracted. To do this, the predicted segmentation is also resampled to meet the dimensions of the new image in world coordinates. Then, the $x$-$y$ coordinates of the ROI are set as a $50 \times 50$ mm box centered around the predicted tumor segmentation. This same 2D ROI is then extracted from (max) 25 slices above and (max) 25 slices below the main slice to generate the 3D $50 \times 50 \times 50$ mm ROI.\footnote{The ``main" slice refers the extracted axial slice upon which the segmentation was originally generated.} For nodules that are at the extremity of the lungs (one of the first or last slices in the series), the number of slices taken from above or below is adjusted if necessary (i.e. if 25 cannot be taken in each direction).

\subsection{Recurrence Prediction Architecture}
\begin{figure*}[!bt]
    \makebox[\textwidth][c]{\includegraphics[width=.8\textwidth]{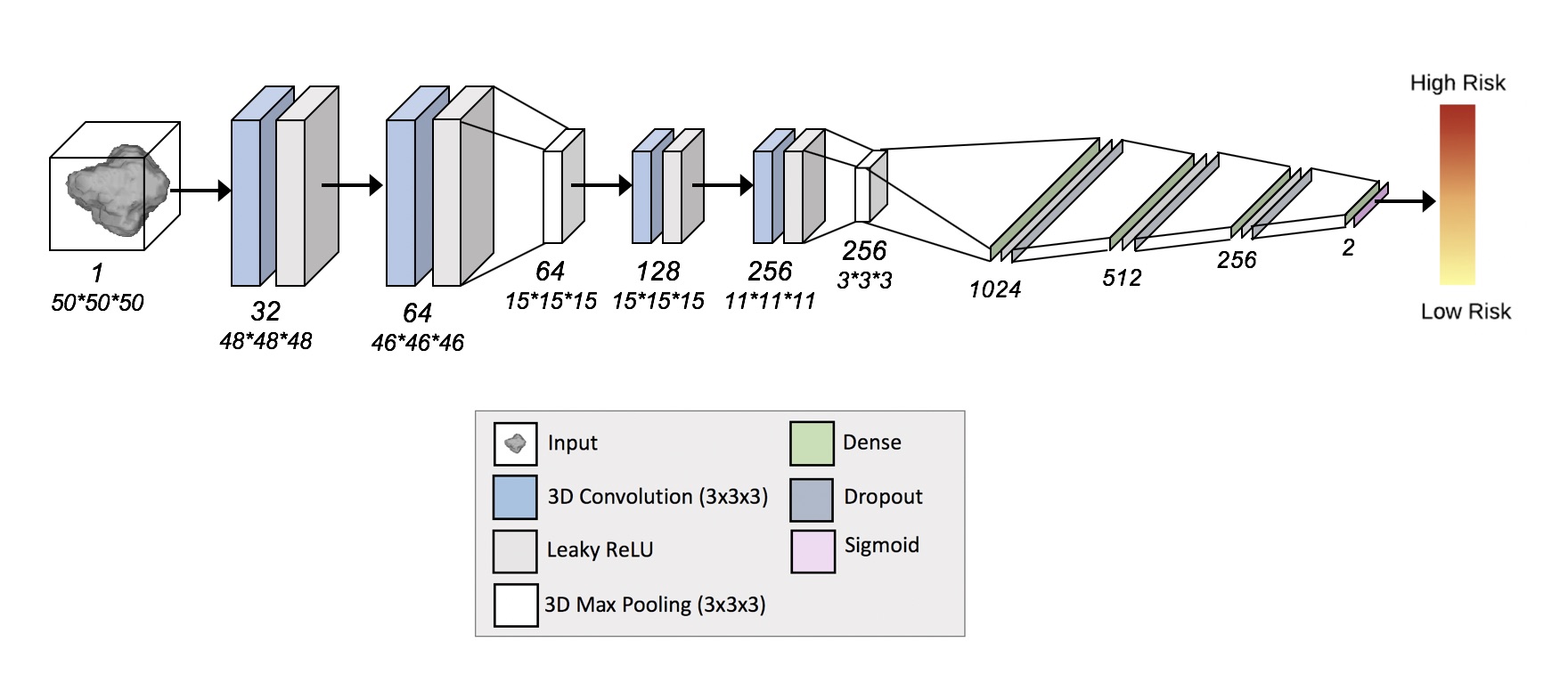}}%
    \caption[3D CNN Architecture for Recurrence Prediction]{3D CNN Architecture for Recurrence Prediction. The output shape (number of channels, then $x,y,z$ dimensions), is denoted below each unit. The colors of the layers represent different operations.}
    \label{fig:stage2_architecture_2}
\end{figure*}

The CNN architecture we use for recurrence prediction is illustrated in Figure \ref{fig:stage2_architecture_2}. It follows a typical CNN structure similar to the one used by Honsy et al. \cite{Hosny2018}. It contains four 3D $(3\times 3\times 3)$ convolutional layers with 32, 64, 128, and 256 filters, respectively. Each convolutional layer is followed by a LeakyReLU activation layer that has an alpha value of 0.1. Additionally, the 2nd and 4th convolutional units are followed by $(3 \times 3 \times 3)$ Max Pooling layers. Following the 4 convolutional units, the images are flattened and fed into 4 fully connected layers with 1,024, 512, 256, and 2 units, respectively. Each fully connected layer has a dropout of 25\%. Finally, a sigmoid activation function is applied to get the final prediction.

\subsection{Recurrence Prediction Training}
Prior to training the network, the images are partitioned into training, testing, and validation sets. Since many patients have multiple nodules, we first create a random split based off patient ids. To ensure an even class distribution, we stratify this split on recurrence status. We then partition the corresponding images for each patient accordingly. Due to the limited number of images in the dataset, data augmentation once again plays a critical role. 

The remaining hyperparameters of our network are similar to those of our segmentation stage. For the optimizer, we use the Adam optimizer with a global learning rate of 0.0001. We also use weighted binary cross-entropy (see Equation \ref{weighted_bce}) for our network's loss function with a weight of 3.0 for the positive class. Our choice is motivated by the disproportional number of patients who do not experience recurrence to those who do experience recurrence.

\subsection{Recurrence Prediction Random Forest}

The final part of our system uses a random forest classifier to combine our neural network's predictions with clinical data and geometric tumor features that are extracted from the segmentation predicted in the previous stage. The model's parameters (number of estimators and maximum tree depth) are selected using cross validation. A list of the features selected is provided below.

\begin{enumerate}
    \item \textit{Clinical Features}: cancer stage, gender, age, adjuvant therapy (whether adjuvant therapy was administered after surgery)
    \item \textit{Geometric features}: diameter, average attenuation, perimeter
    \item \textit{Neural Network Predictions}: in order to go from per nodule predictions to per patient predictions, the ``most confident" predicted probability of any of the patient's nodules is used for each patient. 
\end{enumerate}

\subsection{Recurrence Prediction Results and Experiments}
We use several different metrics to evaluate the performance of our model: 
\begin{enumerate}
    \item \textit{Precision}
    \item \textit{Recall}
    \item \textit{ROC-AUC}
\end{enumerate}
In addition to measuring performance on the overall test set, we also evaluate our model on subgroups of patients based off cancer stage. For stage I patients, we measure accuracy, recall, and for stage II patients, we measure accuracy, recall, and precision. These metrics are chosen to reflect our model's capability to identify (a) high risk stage I patients and (b) low risk stage II patients. As is discussed in our review of the TNM staging system, these are particular prognostic needs that have been expressed in previous medical literature \cite{Subramanian2010}.

\begin{figure*}[!tb]
    \makebox[\textwidth][c]{\includegraphics[width=.75\textwidth]{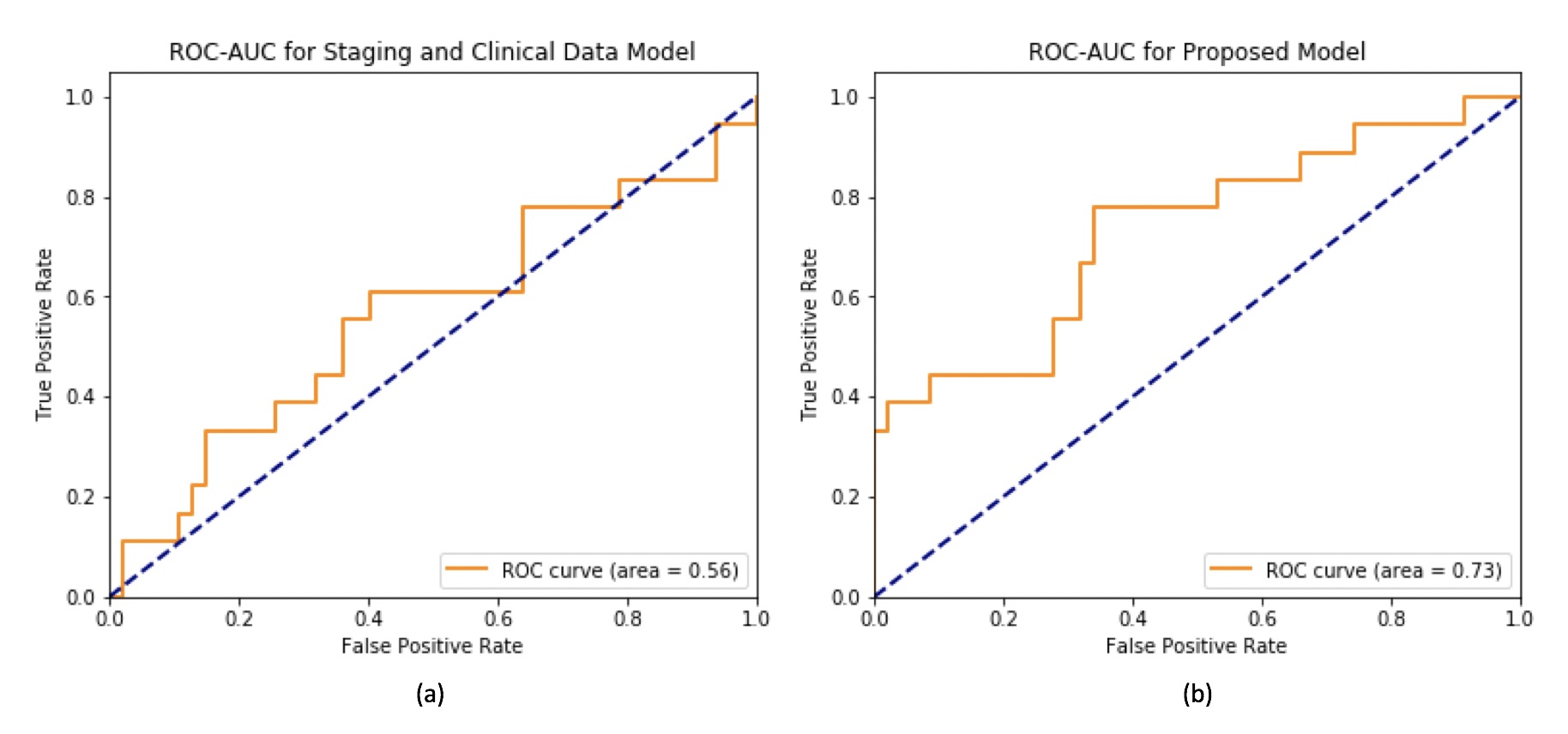}}%
    \caption[Random Forest ROC-AUC for Recurrence Prediction]{Random forest ROC-AUC for recurrence prediction. (a) ROC-AUC of the random forest model using only staging and clinical data. (b) ROC-AUC of the proposed random forest model.}
    \label{fig:rf_roc_auc}
\end{figure*}

We provide the results of our proposed system below. To demonstrate their significance, we also compare the performance of our final model to a random forest that uses only staging and other relevant clinical data (i.e. no geometric or network prediction features). The results from our model and this comparison are shown in Tables \ref{table:rf_overall_recurrence_results} and \ref{table:rf_stage_recurrence_results} and Figure \ref{fig:rf_roc_auc}.

\begin{table}[h!]
\small
\caption[Random Forest Overall Recurrence Prediction Performance]{Random forest overall recurrence prediction performance}
\label{table:rf_overall_recurrence_results}
\centering
\begin{tabular}{||P {2.2cm}|P {1.3cm}| P {.75cm}| P {1.2cm}| P {1.2cm}||} 
 \hline
 \textbf{Model} & Accuracy & AUC & Recall & Precision\\ [0.5ex] 
 \hline\hline
 Staging and Clinical Random Forest Model & 71\% & 56\% & 16\% & 42\%\\ 
 \hline
 Proposed Random Forest Model & 78\% & 73\% & 44\% & 66\%\\ 
 \hline
\end{tabular}
\end{table}

\begin{table}[h!]
\caption[Random Forest Recurrence Prediction Performance by Cancer Stage]
{Random forest recurrence prediction performance by cancer stage}
\label{table:rf_stage_recurrence_results}
\small
\centering
\begin{tabular}{||P {1.7cm}|P {1.3cm}| P {1.2cm}| P {1.25cm}| P {1.2cm}||} 
 \hline
 Method & Stage I Accuracy & Stage I Recall & Stage II Accuracy & Stage II Precision\\ [0.5ex] 
 \hline\hline
 Staging and Clinical Random Forest Model& 75\%   & 0\% & 25\% &0\%\\ 
 \hline
 Proposed Random Forest Model & 75\% & 17\% & 100\% & 100\%\\
 \hline
\end{tabular}
\end{table}

\subsection{Recurrence Prediction Discussion}
The TNM staging system demonstrates the ability to capture a certain, but limited amount of information about a patient's risk for recurrence. It is shown that patients in later stages are more likely to experience recurrent cancer \cite{Consonni2015}. It is also shown from the staging and clinical data model in our experiments that these factors can be utilized to make non-trivial predictions. However, prognostic tools that depend on this data exclusively are also inevitably insufficient. Putting excessive weight on whether or not the patient has progressed to a later stage, has unavoidable limitations. However, the results of our final model demonstrate the efficacy of an integrated approach and more generally, the potential of deep learning and image processing to address these issues. By combining neural network predictions with extracted geometric features and clinical and staging data, the model outperforms all other approaches tested. Both overall performance, and performance within staging subgroups are significantly increased. 
These findings are significant for our study, as the integration of the three different types of data is unique to our approach. The extraction of geometric features from the tumors and use of the extensive clinical data from the NLST dataset is only possible with our fully automated segmentation pipeline.

\section{Conclusion}
In this paper we present a fully automated system for the segmentation and recurrence risk prediction of surgically resected lung tumors. In recent years, deep learning models, and CNNs in particular, have become the state of the art in lung nodule segmentation. Within the realm of recurrence prediction however, these types of approaches have just recently arisen. A significant barrier to the continued development of such models is the lack of viable datasets. Using any sort of image processing method to address recurrence prediction requires a significant amount of very specific features, many of which are not available in standard lung cancer imaging collections. These features include treatment data to determine the procedure used, progression data to determine recurrence status, and usually, data to localize the tumor within the image when no radiologist is available. However, by taking advantage of the progress in nodule segmentation, our proposed system circumvents this need for localization data. To the best of our knowledge, our framework is the first fully automated segmentation and classification system for recurrence risk prediction.

We do acknowledge that the limited size of our datasets and our position outside the medical field are limitations of the study. Nonetheless, we hope that our system will inspire future work that addresses these issues and continues the development of similar approaches \cite{Lambin2017}. While these types of deep learning and radiomics models are novel, they ultimately have the potential to tackle current prognostic challenges and enable a new era of personalized medicine.

\bibliographystyle{unsrt} 
\bibliography{main} 

\begin{thebibliography}{10}

\bibitem{Rahib2014}
L.~Rahib, B.~Smith, R.~Aizenberg, A.~Rosenzweig, J.~Fleshman, and L.~Matrisian,
  ``Projecting cancer incidence and deaths to 2030: the unexpected burden of
  thyroid, liver, and pancreas cancers in the united states,'' {\em Cancer
  Research}, vol.~74, no.~11, pp.~2913--2921, 2014.

\bibitem{Consonni2015}
D.~Consonni, M.~Pierobon, M.~H. Gail, M.~Rubagotti, M.~Rotunno, A.~Goldstein,
  L.~Goldin, J.~Lubin, S.~Wacholder, N.~Caporaso, {\em et~al.}, ``Lung cancer
  prognosis before and after recurrence in a population-based setting,'' {\em
  JNCI: Journal of the National Cancer Institute}, vol.~107, no.~6, 2015.

\bibitem{Sugimura2007}
H.~Sugimura, F.~Nichols, P.~Yang, M.~Allen, S.~Cassivi, C.~Deschamps,
  B.~Williams, and P.~Pairolero, ``Survival after recurrent nonsmall-cell lung
  cancer after complete pulmonary resection,'' {\em The Annals of Thoracic
  Surgery}, vol.~83, no.~2, pp.~409--418, 2007.

\bibitem{Hirsch2017}
F.~Hirsch, G.~Scagliotti, J.~Mulshine, R.~Kwon, W.~Curran~Jr, Y.-L. Wu, and
  L.~Paz-Ares, ``Lung cancer: current therapies and new targeted treatments,''
  {\em The Lancet}, vol.~389, no.~10066, pp.~299--311, 2017.

\bibitem{owidcancer}
M.~Roser and H.~Ritchie, ``Cancer,'' {\em Our World in Data}, 2020.
\newblock Available at \url{https://ourworldindata.org/cancer}.

\bibitem{hayes2014personalized}
D.~F. Hayes, H.~S. Markus, R.~D. Leslie, and E.~J. Topol, ``Personalized
  medicine: risk prediction, targeted therapies and mobile health technology,''
  {\em BMC medicine}, vol.~12, no.~1, p.~37, 2014.

\bibitem{CancerStagingNCI}
``Cancer staging - national cancer institute.'' Available at
  \url{https://www.cancer.gov/about-cancer/diagnosis-staging/staging}.

\bibitem{Osarogiagbon2012}
R.~U. Osarogiagbon, ``Predicting survival of patients with resectable non-small
  cell lung cancer: Beyond {TNM},'' {\em Journal of Thoracic Disease}, vol.~4,
  no.~2, pp.~214--216, 2012.

\bibitem{Hung2009}
J.~Hung, W.~Hsu, C.~Hsieh, B.~Huang, M.~Huang, J.~Liu, and Y.~Wu,
  ``Post-recurrence survival in completely resected stage {I} non-small cell
  lung cancer with local recurrence,'' {\em Thorax}, vol.~64, no.~3,
  pp.~192--196, 2009.

\bibitem{Subramanian2010}
J.~Subramanian and R.~Simon, ``Gene expression--based prognostic signatures in
  lung cancer: ready for clinical use?,'' {\em Journal of the National Cancer
  Institute}, vol.~102, no.~7, pp.~464--474, 2010.

\bibitem{Tsuboi2007}
M.~Tsuboi, T.~Ohira, H.~Saji, K.~Miyajima, N.~Kajiwara, O.~Uchida, J.~Usuda,
  and H.~Kato, ``The present status of postoperative adjuvant chemotherapy for
  completely resected non-small cell lung cancer,'' {\em Annals of Thoracic and
  Cardiovascular Surgery}, vol.~13, no.~2, pp.~73--77, 2007.

\bibitem{JungKoo2017}
H.~J. Koo, Y.~S. Sung, W.~H. Shim, H.~Xu, C.~M. Choi, H.~R. Kim, J.~B. Lee, and
  M.~Y. Kim, ``Quantitative computed tomography features for predicting tumor
  recurrence in patients with surgically resected adenocarcinoma of the lung,''
  {\em PLoS One}, vol.~12, no.~1, 2017.

\bibitem{Hosny2018}
A.~Hosny, C.~Parmar, T.~Coroller, P.~Grossmann, R.~Zeleznik, A.~Kumar,
  J.~Bussink, R.~Gillies, R.~Mak, and H.~J. Aerts, ``Deep learning for lung
  cancer prognostication: A retrospective multi-cohort radiomics study,'' {\em
  PLoS Medicine}, vol.~15, no.~11, p.~e1002711, 2018.

\bibitem{Rizzo2018}
S.~Rizzo, F.~Botta, S.~Raimondi, D.~Origgi, C.~Fanciullo, A.~G. Morganti, and
  M.~Bellomi, ``Radiomics: the facts and the challenges of image analysis,''
  {\em European Radiology Experimental}, vol.~2, no.~1, pp.~1--8, 2018.

\bibitem{Lambin2012}
P.~Lambin, E.~Rios-Velazquez, R.~Leijenaar, S.~Carvalho, R.~Van~Stiphout,
  P.~Granton, C.~Zegers, R.~Gillies, R.~Boellard, A.~Dekker, {\em et~al.},
  ``Radiomics: extracting more information from medical images using advanced
  feature analysis,'' {\em European Journal of Cancer}, vol.~48, no.~4,
  pp.~441--446, 2012.

\bibitem{Aberle2011}
{National Lung Screening Trial Research Team}, ``Reduced lung-cancer mortality
  with low-dose computed tomographic screening,'' {\em New England Journal of
  Medicine}, vol.~365, no.~5, pp.~395--409, 2011.

\bibitem{Kazerooni2015}
E.~Kazerooni, M.~Armstrong, J.~Amorosa, D.~Hernandez, L.~Liebscher, H.~Nath,
  M.~McNitt-Gray, E.~Stern, and P.~Wilcox, ``{ACR} {CT} accreditation program
  and the lung cancer screening program designation,'' {\em Journal of the
  American College of Radiology}, vol.~13, pp.~38--42, Feb. 2016.

\bibitem{Moyer2014}
V.~A. Moyer, ``{Screening for lung cancer: U.S. preventive services task force
  recommendation statement},'' {\em Annals of Internal Medicine}, vol.~160,
  pp.~330--338, Mar. 2014.

\bibitem{Wood2018}
D.~Wood, E.~Kazerooni, S.~Baum, G.~Eapen, D.~Ettinger, L.~Hou, D.~Jackman, {\em
  et~al.}, ``Lung cancer screening, version 3.2018,'' {\em JNCCN Journal of the
  National Comprehensive Cancer Network}, vol.~16, pp.~412--441, Apr. 2018.

\bibitem{Armato2011}
S.~Armato, G.~McLennan, L.~Bidaut, M.~McNitt-Gray, C.~Meyer, A.~Reeves,
  B.~Zhao, {\em et~al.}, ``The lung image database consortium {(LIDC)} and
  image database resource initiative {(IDRI)}: A completed reference database
  of lung nodules on {CT} scans,'' {\em Medical Physics}, vol.~38,
  pp.~915--931, Jan. 2011.

\bibitem{Cavalcanti2016}
P.~Cavalcanti, S.~Shirani, J.~Scharcanski, C.~Fong, J.~Meng, J.~Castelli, and
  D.~Koff, ``Lung nodule segmentation in chest computed tomography using a
  novel background estimation method,'' {\em Quantitative Imaging in Medicine
  and Surgery}, vol.~6, no.~1, pp.~16--24, 2016.

\bibitem{Wang2017}
S.~Wang, M.~Zhou, Z.~Liu, Z.~Liu, D.~Gu, Y.~Zang, D.~Dong, O.~Gevaert, and
  J.~Tian, ``Central focused convolutional neural networks: Developing a
  data-driven model for lung nodule segmentation,'' {\em Medical Image
  Analysis}, vol.~40, pp.~172--183, Aug. 2017.

\bibitem{Wang2019}
W.~Wang, R.~Feng, J.~Chen, Y.~Lu, T.~Chen, H.~Yu, D.~Chen, and J.~Wu,
  ``Nodule-plus {R-CNN} and deep self-paced active learning for {3D} instance
  segmentation of pulmonary nodules,'' {\em IEEE Access}, vol.~7,
  pp.~128796--128805, Sept. 2019.

\bibitem{Nithila2016}
E.~Nithila and S.~Kumar, ``Segmentation of lung nodule in {CT} data using
  active contour model and fuzzy c-mean clustering,'' {\em Alexandria
  Engineering Journal}, vol.~55, pp.~2583--2588, Sept. 2016.

\bibitem{liaw2002classification}
A.~Liaw, M.~Wiener, {\em et~al.}, ``Classification and regression by
  randomforest,'' {\em R news}, vol.~2, no.~3, pp.~18--22, 2002.

\bibitem{nlst}
{National Lung Screening Trial Research Team}, ``The national lung screening
  trial: overview and study design,'' {\em Radiology}, vol.~258, no.~1,
  pp.~243--253, 2011.

\bibitem{Ronneberger}
O.~Ronneberger, P.~Fischer, and T.~Brox, ``U-net: Convolutional networks for
  biomedical image segmentation,'' in {\em Medical Image Computing and
  Computer-Assisted Intervention (MICCAI)}, vol.~9351 of {\em LNCS},
  pp.~234--241, Springer, 2015.

\bibitem{Zhang}
Z.~Zhang, Q.~Liu, and Y.~Wang, ``Road extraction by deep residual u-net,'' {\em
  IEEE Geoscience and Remote Sensing Letters}, vol.~15, no.~5, pp.~749--753,
  2018.

\bibitem{He}
K.~He, X.~Zhang, S.~Ren, and J.~Sun, ``Deep residual learning for image
  recognition,'' in {\em Proceedings of the IEEE conference on computer vision
  and pattern recognition}, pp.~770--778, 2016.

\bibitem{Kamal}
U.~Kamal, A.~Rafi, R.~Hoque, and M.~Hasan, ``Lung cancer tumor region
  segmentation using recurrent {3D-DenseUNet},'' {\em arXiv preprint
  arXiv:1812.01951}, 2018.

\bibitem{Lambin2017}
P.~Lambin, R.~Leijenaar, T.~Deist, J.~Peerlings, E.~De~Jong, J.~Van~Timmeren,
  S.~Sanduleanu, R.~Larue, A.~Even, A.~Jochems, {\em et~al.}, ``Radiomics: the
  bridge between medical imaging and personalized medicine,'' {\em Nature
  Reviews Clinical Oncology}, vol.~14, no.~12, p.~749, 2017.

\end{thebibliography}

\end{document}